%% file: main.tex
\pdfoutput=1
\documentclass[11pt]{article}

\usepackage[]{acl}

\usepackage{times}
\usepackage{latexsym}
\usepackage[T1]{fontenc}
\usepackage[utf8]{inputenc}
\usepackage{microtype}
\usepackage{inconsolata}

\usepackage{algorithm}
\usepackage{algorithmic}
\usepackage{subcaption}
\usepackage{booktabs}
\usepackage{multirow}
\usepackage{tabularx} 
\usepackage{comment}
\usepackage{xspace}
\usepackage{makecell}
\usepackage{mathtools}
\usepackage{enumitem}
\usepackage{tikz}
\usepackage{textcomp}
\usepackage{amsmath,amsfonts}
\usepackage[noabbrev,capitalize]{cleveref}
\usepackage{url}
\usepackage{newfloat}
\usepackage{listings}
\usepackage{bibentry}
\usepackage{dsfont}
\usepackage{listings}
\usepackage{fancyhdr}

\definecolor{codegreen}{rgb}{0,0.6,0}
\definecolor{codegray}{rgb}{0.5,0.5,0.5}
\definecolor{codepurple}{rgb}{0.58,0,0.82}
\definecolor{backcolour}{rgb}{0.95,0.95,0.92}
\definecolor{forestgreen}{rgb}{0.24, 0.71, 0.54}
\definecolor{goldenyellow}{rgb}{0.85, 0.57, 0.0}
\definecolor{internationalkleinblue}{rgb}{0.0, 0.47, 0.75}

\lstdefinestyle{python}{
language=Python,
backgroundcolor=\color{backcolour},   
commentstyle=\color{codegreen},
keywordstyle=\color{magenta},
numberstyle=\tiny\color{codegray},
stringstyle=\color{codepurple},
basicstyle=\ttfamily\scriptsize,
escapeinside={<@}{@>},
breakatwhitespace=false,         
breaklines=true,                 
keepspaces=true,                 
numbers=left,       
numbersep=5pt,                  
showspaces=false,                
showstringspaces=false,
showtabs=false,                  
tabsize=1,
frame=c,
framesep=4.5mm,
framexleftmargin=4mm,
xleftmargin=.05\linewidth, 
}

\lstnewenvironment{terminal}{%
  \thicklines
  \lstset{
    backgroundcolor=\color{backcolour},
    rulecolor=\color{backcolour!40!black},
    frameround=tttt,
    frame=single,
    basicstyle=\ttfamily\scriptsize,
    basewidth=0.50em,
    keywordstyle=\bfseries,
    xleftmargin=.02\linewidth,
    xrightmargin=.02\linewidth,
    escapeinside={<@}{@>},
  }}{}

\lstnewenvironment{python}{%
  \thicklines
  \lstset{
    language=Python,
    commentstyle=\color{codegreen},
    keywordstyle=\color{magenta},
    numberstyle=\tiny\color{codegray},
    stringstyle=\color{codepurple},
    backgroundcolor=\color{backcolour},
    rulecolor=\color{backcolour!40!black},
    frameround=tttt,
    frame=single,
    basicstyle=\ttfamily\scriptsize,
    basewidth=0.50em,
    xleftmargin=.02\linewidth,
    xrightmargin=.02\linewidth,
    escapeinside={<@}{@>},
    breakatwhitespace=false,         
    breaklines=true,                 
    keepspaces=true,                                  
    showspaces=false,                
    showstringspaces=false,
    showtabs=false,                  
    tabsize=1,
  }}{}

\newcommand{\ie}{\mbox{i.e.}\xspace}
\newcommand{\eg}{\mbox{e.g.}\xspace}
\newcommand{\our}{\mbox{AgentQuest}\xspace}

\title{\our: Benchmarking LLM Agents Behaviours in \\ Multi-step Intensive Reasoning Tasks}
\title{\our: A Modular Benchmark Framework \\ to Measure Progress and Improve LLM Agents}

 \author{Luca Gioacchini$^{1,2}$, Giuseppe Siracusano$^1$, Davide Sanvito$^1$, Kiril Gashteovski$^{1,3}$, \\\textbf{David Friede$^1$, Roberto Bifulco$^1$, Carolin Lawrence$^1$}\\
   $^1$ NEC Laboratories Europe, Heidelberg, Germany  \\
   $^2$ Politecnico di Torino, Turin, Italy  \\
   $^3$ CAIR, Ss. Cyril and Methodius University, Skopje, North Macedonia}

\begin{document}
\maketitle

\begin{abstract}
The advances made by Large Language Models (LLMs) have led to the pursuit of LLM agents that can solve intricate, multi-step reasoning tasks. As with any research pursuit, benchmarking and evaluation are key corner stones to efficient and reliable progress. However, existing benchmarks are often narrow and simply compute overall task success. To face these issues, we propose \our\footnote{Demo provided at \url{https://youtu.be/0JNkIfwnoak}.} -- a framework where (i) both benchmarks and metrics are modular and easily extensible through well documented and easy-to-use APIs; (ii) we offer two new evaluation metrics that can reliably track LLM agent progress while solving a task. We exemplify the utility of the metrics on two use cases wherein we identify common failure points and refine the agent architecture to obtain a significant performance increase. Together with the research community, we hope to extend \our further and therefore we make it available under \url{https://github.com/nec-research/agentquest}.

\end{abstract}

\input{sections/intro}
\input{sections/ai_agents}
\input{sections/frame_howto}
\input{sections/insights}
\input{sections/conclusions}

\section*{Acknowledgements}
This project has received funding from the European Union’s Horizon Europe research and innovation programme (SNS-JU) under the Grant Agreement No 101139285 (``NATWORK”).

\bibliography{bibliography}
\clearpage
\appendix
\input{appendix/app1}

\end{document}

%% file: sections/intro.tex
\section{Introduction}
\label{s:intro}

Generative Agents~\cite{kiela2023saturation} are software systems that leverage foundation models like Large Language Models (LLMs) to perform complex tasks, take decisions, devise multi-steps plans and use tools (API calls, coding, etc.) to build solutions in heterogeneous contexts~\cite{wang2023survey,weng2023prompt}.
The potential ability to solve heterogeneous tasks with high degrees of autonomy has catalysed the interest of both research and industrial communities. Nonetheless, it is still unclear to which extent current systems are successfully able to fulfil their promises. In fact, methodologies to benchmark, evaluate and advance these systems are still in their early days.

\begin{figure}[!t]
    \centering
    \begin{subfigure}[b]{.48\linewidth}
        \centering
        \includegraphics[width=\linewidth]{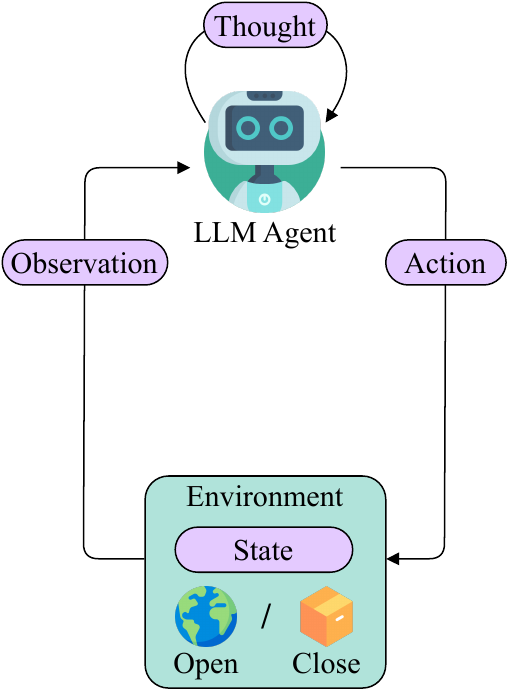}
        \caption{Existing}
        \label{fig:interaction_before}
    \end{subfigure}
    \hspace{.3em}
    \begin{subfigure}[b]{.48\linewidth}
        \centering
        \includegraphics[width=\linewidth]{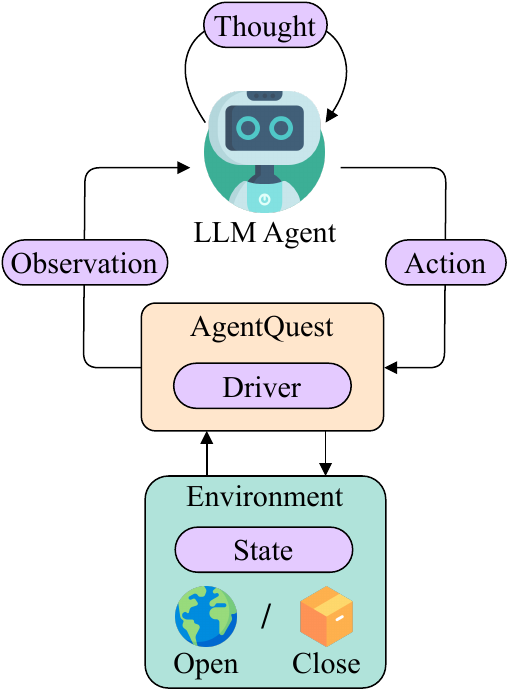}
        \caption{\our}
        \label{fig:interaction_after}
    \end{subfigure}
    \caption{Overview of agent-benchmark interactions in existing frameworks and in \our. \our defines a common interface to interact with the benchmarks and to compute progress metrics, easing the addition of new benchmarks and allowing researchers to evaluate and debug their agent architectures.}
    \label{fig:interaction}
\end{figure}

We identify a couple of gaps. 
Firstly, benchmarking agents requires combining  different benchmark types~\cite{liu2023agentbench,chalamalasetti2023clembench}. For example, some benchmarks focus on specific capabilities and provide gaming environments, which we refer to as ``closed-box'' -- \ie with a finite set of actions~\cite{liu2023agentbench, patil2023gorilla,chalamalasetti2023clembench} -- whereas other benchmarks provide open-ended tasks and access to general tools, like web browsing~\cite{zhuang2023toolqa, zheng2023judging,mialon2023gaia}. As benchmarks are developed independently, significant effort goes into custom integration of new agent architectures with each benchmark.

Secondly, and more critically, existing benchmarks mostly focus on providing a \emph{success rate} measure, \ie a binary success/fail evaluation for each of the proposed tasks. 
While success rate is helpful to measure overall advances of an agent technology, it has limited use in guiding improvements for new generative agent architectures. Here, it is important to consider that generative agents often combine foundation models with multiple other components, such as memory and tools. Developers can reason about these individual components in terms of architecture and their inter-dependence, and could actively change and evolve them using deeper insights about how an agent performs in a benchmark. That is, developers need benchmarks to both evaluate and \emph{debug} agents.

For example, current benchmarks make it hard to answer questions like \textit{does the agent fail completely the tasks or does it partially solve them?} \textit{Does the agent fail consistently at a certain step?} \textit{Would extra run time lead to finding a solution?}
Answering these questions would require tracing and inspecting the execution of the agent.
We argue that providing a more efficient approach that is consistent over multiple benchmarks is a stepping stone towards evolving generative agents.

We address these gaps introducing \our, a modular framework to support multiple diverse benchmarks and agent architectures (See Figure~\ref{fig:interaction}), alongside with two new metrics -- \ie progress rate and repetition rate -- to debug an agent architecture behaviour. 
\our defines a standard interface to connect an arbitrary agent architecture with diverse benchmarks, and to compute progress and repetition rates from them. 

We showcase the framework, implementing 4 benchmarks in \our: ALFWorld~\cite{shridhar2020alfworld}, Lateral Thinking Puzzles~\cite{sloane1992lateral}, Mastermind and Sudoku. The latter two are newly introduced with \our.
Additional benchmarks can be easily added, while requiring no changes to the tested agents.

Our final contribution is to present our experience leveraging the proposed metrics to debug and improve existing agent architectures as implemented in LangChain~\cite{chase2022langchain}. In particular, we show that in the Mastermind benchmark the combination of progress rate and repetition rate identifies a limitation in the ability of the agent to explore the full space of potential solutions. Guided by this insight we could improve the success rate in this benchmark by up to $\approx$20\%.
In Lateral Thinking Puzzles we show that partially repeating actions is part of the agent strategy, whereas in ALFWorld, we show that monitoring the progress rate makes it possible to identify that the final success rate is limited by the allowed runtime of the agent, and that more steps lead to a better performance. 
Finally, in the Sudoku benchmark, we show that the low success rate is actually paired with low progress rate, making clear that the tested agent is unable to solve this type of tasks.

%% file: sections/ai_agents.tex
\section{Generative AI Agents in a Nutshell}
\label{s:framework}
Generative AI agents are automated systems relying on software components integrated with LLMs pre-trained on large amount of data for language understanding and processing. 
When assigned a task, an agent engages in a systematic process: it iteratively formulates self-generated instructions, executes them, and observes the outcomes until the ultimate objective is achieved.
Next, we showcase the basic interaction between agents and the environment in which they operate and describe the standard benchmarking techniques.

\subsection{Agent-Environment interaction}
Closely following the terminology in Reinforcement Learning (RL)\footnote{Unlike RL scenarios, the agent does not need a further training process. It  relies on the pre-trained LLM and does not perform an action under the influence of any reward.}~\cite{sutton2018reinforcement}, the core elements defining the agent-environment interaction are \emph{environment}, \emph{state}, \emph{observation} and \emph{action} (see \cref{fig:interaction_before}).

\paragraph{Environment and states.}
The environment refers to the external system the agent interacts with. In this context, we treat the benchmark and the environment as synonyms. It is typically described through a finite set of hidden \emph{states}, which are not directly observable by the agent and represent the benchmark configuration.

\paragraph{Observations and actions.}
The agent interacts with the environment for multiple execution steps. At each step, the environment produces an \emph{observation} providing information about its current hidden state. The agent uses the internal LLM to process the received observation. 
Being pre-trained on general knowledge data, the LLM engages a reasoning process generating a \emph{thought} on the observation (\eg the planned strategy to follow in the current step or the usage of a tool). According to this thought, the agent provides the environment an \emph{action} to modify the current hidden state.
\footnote{Unlike RL, the LLM outputs are unconstrained, and any provided action is considered valid.}

The following is an example of an agent interacting with a closed-box environment to solve a code-breaking challenge, \ie the Mastermind game~\cite{stuckman2005mastermind}. 
The agent is tasked to guess a four-digit code. At each step, the agent devises a strategy and acts accordingly to guess the code. The environment provides a feedback on the number of guessed digits in correct and wrong positions. The process continues until a stopping condition is met.

\begin{terminal}
<@\textcolor{forestgreen}{\textbf{ENVIRONMENT}: Initializing a new environment...}@>
<@\textcolor{internationalkleinblue}{\textbf{OBSERVATION}: Start guessing the 4 digits code.}@>
<@\textcolor{magenta}{\textbf{THOUGHT}: I'll start by guessing 1234 to get an idea of which}@>
<@\textcolor{magenta}{of these numbers are in the secret number and if any are in}@>
<@\textcolor{magenta}{the correct position.}@>
<@\textcolor{goldenyellow}{\textbf{ACTION}: 1234}@>

<@\textcolor{forestgreen}{\textbf{ENVIRONMENT}: Comparing the guessed code with the truth...}@>
<@\textcolor{internationalkleinblue}{\textbf{OBSERVATION}: Your guess has 1 correct numbers in the wrong }@>
<@\textcolor{internationalkleinblue}{position and 0 correct numbers in the correct position. }@>
<@\textcolor{internationalkleinblue}{Keep guessing...}@>
<@\textcolor{magenta}{\textbf{THOUGHT}: One of the digits 1, 2, 3, or 4 is correct but in}@>
<@\textcolor{magenta}{the wrong position. I'll try to move each digit to a}@>
<@\textcolor{magenta}{different position to see if I can get a correct position.}@>
<@\textcolor{goldenyellow}{\textbf{ACTION}: 2143}@>
...
\end{terminal}

\subsection{Benchmarking an agent}
When evaluating agents performance on a benchmark, the following two metrics are commonly used~\cite{liu2023agentbench}: (i) Success Rate (SR), \ie the ratio of successful tasks to the total instances; (ii) Time to Success, \ie the average time required to obtain a solution. While important and trending metrics~\cite{chalamalasetti2023clembench,hessel2022clipscore,zhang2020bertscore}, they exclusively address the final success. They cannot measure intermediate success or failure and therefore make it difficult to understand why agents might systematically fail and how they can be improved. 
In contrast, we want to define intermediate metrics that allow us to easily assess and compare the performance of agents across a wide range of tasks.

%% file: sections/frame_howto.tex
\section{\our Overview}
\label{s:howto}
We designed \our as a separation layer between agent and environment (see~\cref{fig:interaction_after}). Essentially, it offers (i) a unified interface (\ie the \emph{driver}) ensuring compatibility between different agent architectures and benchmarks with minimal programming efforts (Section \ref{subsec:driver}); (ii) the implementation of two metrics beyond task success (\ie \emph{progress rate} and \emph{repetition rate}) aimed at monitoring the  agent advancement toward the final goal and allowing us to understand the reasons behind failures (Section \ref{subsec:metrics}); (iii) a unique vantage point and interface for implementing new metrics to monitoring and measuring the execution (Section~\ref{subsec:new_metrics}).

\subsection{Benchmarks common interface}\label{subsec:driver}
Different benchmarks require invoking distinct functions, using specific formats, and performing parsing and post-processing of observations and agent actions. To integrate different agent architectures, the common trend is hardcoding such benchmark-specific requirements directly in the framework~(\citealt{liu2023agentbench,chalamalasetti2023clembench}, \textit{inter alia}). This results in many custom interfaces tailored on each environment, making it difficult to easily move to other benchmarks and agent architectures.

Instead, \our exposes a single unified Python interface, \ie the \texttt{Driver} and two classes reflecting the agent-environment interaction components (\ie \texttt{Observation}, \texttt{Action}). 

\paragraph{Observations and actions.} We provide two simple classes: \texttt{Observation} and \texttt{Action}. The first has two required attributes: (i) \texttt{output}, a string reporting information about the environment state; (ii) \texttt{done}, a Boolean variable indicating if the final task is currently accomplished or not.
The \texttt{Action} class has one required attribute, \texttt{action\_value}. It is a string directly output by the agent. Once processed and provided to the environment, it triggers the environment change. To customise the interactions, developers can define optional attributes.

\paragraph{Driver.}
We provide the \texttt{Driver} class with two mandatory methods: (i) the \texttt{reset} method initialises a new instance of the environment and returns the first observation; (ii) the \texttt{step} method performs one single execution step. It accepts one instance of the \texttt{Action} class from the agent, processes the action (\eg parses the \texttt{action\_value} string) and uses it to modify the environment state. 
It always returns an observation. 
The driver supports also the benchmark-specific \texttt{state} attribute, acting as a simple API. It exposes the environment state at step $t$, useful to compute the progress rate.

We here provide an example of the implemented interaction for Mastermind:

\begin{python}
from agentquest.drivers import MasterMindDriver
from agentquest.utils import Action
from agentquest.metrics import get_progress, get_repetition

agent = ... # Initialize your agent
actions, progress, repetitions = [], [], []
# Initialize the environment and reset round
driver = MasterMindDriver(truth='5618')
obs = driver.reset()
# Agent loop
while not obs.done:
    guess = agent(obs.output) # Get the agent output
    action = Action(action_value=guess) # Create action
    actions.append(action.action_value) # Store action
    obs = driver.step(action) # Execute step
    # Compute current progress and repetition
    progress.append(get_progress(driver.state, '5618'))
    repetitions.append(get_repetitions(actions))
    # Extend with your custom metrics here ...
# Compute final metrics
PR = [x/len('5618') for x in progress]
RR = [x/(len(actions)-1) for x in repetitions]
\end{python}

\subsection{Understanding agent advancements}
\label{subsec:metrics}

Getting insights on how they tackle a specific task is key to comprehend agent behaviours, capabilities and limitations. Furthermore, identifying systematic agent failures allows to pinpoint necessary adjustments within the architecture to effectively address the underlying issues.

\our contributes towards this direction introducing two cross-benchmark metrics, the \emph{progress rate} and the \emph{repetition rate}. While the first expresses \emph{how much} the agent is advancing towards the final goal, the latter indicates \emph{how} it is reaching it, with a specific focus on the amount of repeated (\ie similar) actions the agent performs.

\paragraph{Milestones and progress rate.} 
To quantify the agent advancement towards the final goal, \our uses a set of \emph{milestones} $\mathcal{M}$. In a nutshell, we break down the final solution into a series of environment hidden states the agent needs to reach to get the final solution of the task, hence, $\mathcal{M} \subseteq \mathcal{S}$, where $\mathcal{S}$ is the set of hidden states. The magnitude of $\mathcal{M}$ determines the level of \emph{granularity} in the evaluation process. Specifically, when $\mathcal{M}$ aligns closely with $\mathcal{S}$, it offers a more comprehensive insight into the agent progress, resulting in finer granularity, whereas for $|\mathcal{M}|=1$ the evaluation coincides with the success rate.

We assign a score to all the states included in $\mathcal{M}$ through a scoring function $f$ and, at execution step $t$, we define the \emph{progress rate} $\text{PR}_t:\mathcal{S}\to [0, 1]$ dependant of such scoring function, as an indication of how far the agent is from the goal, allowing to track agent progress over time. Depending on the benchmark, the progress rate might also decrease during the execution. Milestones can either be manually annotated, or internally computed.

\paragraph{Repetition rate.} 
The repetition rate $\text{RR}_t$ is a measure of the agent tendency of repeating actions. Depending on the benchmark, we do not consider repetitions as a limitation 
-- \eg solving a maze requires repetitions, such as going left repeatedly. See also \cref{s:results} for a positive and negative example of repetitions.

At execution step $t$, we consider the set of unique actions taken by the agent up to $t-1$, $\mathcal{A}_{t-1}$. Then, we compute the similarity function $g$ between the current action $a_t$ and all the previous ones in $\mathcal{A}_{t-1}$. 
As any action generated by the LLM is considered valid, we consider the action $a_t$ as \emph{repeated} if it exists at least one previous action $a \in \mathcal{A}_{t-1}$ such that $g(a_t, a)\geq\theta_a$, where $\theta_a \in [0, 1]$ is the \emph{resolution}.\footnote{A higher resolution demands closer matches for classification as repeated actions, while lower values broaden the spectrum of qualifying action similarities.}
If the action is not repeated, we update the set of unique actions as $A_t = A_{t-1} \cup a_t$.

Based on this, we define the repetition rate at step $t$ as the cumulative number of repeated actions normalised by the number of execution steps, $T$, except for the first.  
Formally, $\text{RR}_t = \frac{t-|A_t|}{T-1}$.

\subsection{Adding new metrics}
\label{subsec:new_metrics}

\begin{table}[]
\small
\centering
\caption{Attributes exposing components of the agent-environment interaction useful to define new metrics.}
\begin{tabular}{lll}
\toprule
\textbf{Class} & \textbf{Attribute} & \textbf{Access to} \\
\midrule
\texttt{Driver} & \texttt{state} & Hidden states \\
\texttt{Observation} & \texttt{output} & Observations \\
\texttt{Action} & \texttt{action$\_$value} & Agent actions \\
\bottomrule
\end{tabular}
\label{tab:recap}
\end{table}

We rely on the progress and repetition rates to show how \our can be extended with new metrics through a simple function template. We then show the implementations of the functions adapted to the considered benchmark.

\paragraph{Metric function template.}

We use a Python function template to easily define the elements of the agent-environment interactions required for computing a given metric. 
\cref{tab:recap} provides a recap of the main attributes and reference classes that can be used as input for the custom metrics. Additionally, users can provide external data, like milestones or action history.

\begin{table*}[t]
\centering
\small
\caption{Overview of the benchmarks provided in \our.}
\begin{tabular}{lll}
\toprule
\textbf{Benchmark} & \textbf{Description} & \textbf{Milestones}\\
\midrule
Mastermind & \begin{tabular}[c]{@{}l@{}}Guessing a numeric code with feedback on guessed digits and positions.\end{tabular} & \begin{tabular}[c]{@{}l@{}}Digits of the code to guess.\end{tabular}\\
\midrule
LTP & \begin{tabular}[c]{@{}l@{}}Solving riddles by asking Yes/No questions.\end{tabular} & \begin{tabular}[c]{@{}l@{}}Guessed riddle key aspects.\end{tabular}\\
\midrule
ALFWorld & \begin{tabular}[c]{@{}l@{}}Finding an object in a textual world and using it.\end{tabular} & \begin{tabular}[c]{@{}l@{}}Sequence of actions.\end{tabular}\\
\midrule
Sudoku & \begin{tabular}[c]{@{}l@{}}9x9 grid puzzle. Digits 1-9 fill each column, row, and 3x3 sub-grid \\without repetition.\end{tabular} & \begin{tabular}[c]{@{}l@{}}Total number of correct \\ inserted digits.\end{tabular}\\
\bottomrule
\end{tabular}
\label{tab:benchmarks}
\end{table*}

\paragraph{Implement progress rate.}
Depending on the benchmark, developers need to implement the custom scoring function $f$ through the \texttt{get\_progress} function and define the set of milestones $\mathcal{M}$. Milestones can either be user-defined or internally computed within \texttt{get\_progress}. Here, we show the definition of \texttt{get\_progress} to quantify the achieved milestones for Mastermind. The milestones are the digits of the final solution and the progress indicates the count of correctly guessed digits in their positions:

\begin{python}
def get_progress(state, milestones):
    reached_milestones = 0 # Digits in correct position
    for i, j in zip(state, milestones):
        if i == j: reached_milestones += 1
    return reached_milestones

# Usage example. The code to guess is '5618'
progress = get_progress('2318', '5618') # Reached milestones
>>> 2
progress/len('5618') # Compute Progress Rate
>>> 0.5
\end{python}

\paragraph{Implement repetition rate.}
To determine if an action is repeated, the end user must define the similarity function $g$ according to the considered benchmark. We provide the \texttt{get\_repetitions} template function to compute the number of repeated actions. Here, we illustrate its implementation in Python and provide a usage example for Mastermind, where $g$ is the Levenshtein similarity~\cite{levenshtein1966binary}.

\begin{python}
from Levenshtein import ratio as g

def get_repetitions(actions, THETA_A):
    unique_act = set() # Initialise unique actions
    for i,a in enumerate(actions):
        # Check for repetitions
        if all([g(a,actions[x])<THETA_A for x in range(i)]):
            unique_act.add(a)
    return len(actions)-len(unique_act)

# Usage example. The code to guess is '5618'
actions = ['1234', '2143', '1234', '5618'] # Actions history
repetitions = get_repetitions(actions, 1.0) 
>>> 1 repeated action
# Compute Repetition Rate
repetitions/(len(actions)-1)
>>> 0.33
\end{python}

In other cases, where $a$ can be any text string, we can use standard metrics, such as BLEU~\cite{bleu}, ROUGE~\cite{rouge} or BERTScore~\cite{bertscore}.

%% file: sections/insights.tex
\section{Insights via \our}
\label{s:results}

We investigate agent behaviours in different reasoning scenarios by proposing a starting set of four benchmarks. We implemented from scratch Sudoku~\cite{felgenhauer2006mathematics} and Mastermind~\cite{stuckman2005mastermind} environments, while ALFWorld~\cite{shridhar2020alfworld} and Lateral Thinking Puzzles (LTP)\cite{sloane1992lateral} are existing implementations~\cite{liu2023agentbench}. 
\cref{tab:benchmarks} provides an overview of the benchmarks and their respective milestones used to measure progress. 

We emphasise that this evaluation is not aimed at providing a thorough evaluation and comparison of agent architectures, but rather to show how to use \our and how monitoring progress and action repetition can provide relevant insights to developers, even after a few executions.

\paragraph{Experimental setup.} 
We use as reference architecture the off-the-shelf chat agent provided by LangChain~\cite{chase2022langchain} powered by GPT-4~\cite{openai2023gpt4} as LLM because it is intuitive, easy to extend and open source.
We run 15 instances of the four benchmarks within \our, setting the maximum number of execution steps as 60\footnote{We limit the number of instances in our experiments for two main reasons: (i) the work primarily serves as a demonstration of the developed framework itself, rather than an extensive evaluation of the agent performance; (ii) extensive tests could have significantly impacted the ability to reproduce the experiments due to the expensive nature of API calls.}. 
In Appendix~\ref{sec:plugandplay} we provide examples on how to use \our with two additional agent architectures and GAIA~\cite{mialon2023gaia} as open-ended environment.

\begin{table}[!t]
\small
\centering
\caption{Average existing and proposed metrics for the tested benchmarks. We report the metrics, Success Rate (SR), Steps, Progress Rate at step 60 (PR$_{60}$) and Repetition Rate at final step 60 (RR$_{60}$). We denote with $^*$ the improved results after modifying the agent architecture.}
\begin{tabular}{l|cc|cc}
\toprule
& \multicolumn{2}{c|}{\textbf{Existing Metrics}} & \multicolumn{2}{c}{\textbf{\our}}\\
& \textbf{SR} & {\textbf{Steps}} & \textbf{PR}$_\mathbf{{60}}$ & \textbf{RR}$_\mathbf{{60}}$ \\
\midrule
Mastermind & 0.47 & 41.87 & 0.62 & 0.32 \\
LTP & 0.20 & 52.00 & 0.46 & 0.81 \\
ALFWorld & 0.86 & 21.00 & 0.74 & 0.06 \\
Sudoku & 0.00 & 59.67 & 0.08 & 0.22 \\
\midrule
Mastermind$^*$ & 0.60 & 39.73 & 0.73 & 0.00 \\
ALFWorld$^*$ & 0.93 & 25.86 & 0.80$^\dagger$ & 0.07$^\dagger$ \\
\bottomrule
\multicolumn{5}{l}{\begin{tabular}[c]{@{}l@{}}$^\dagger$Metrics referred to the extended runtime up to 120\\steps, hence PR$_{120}$ and RR$_{120}$.\end{tabular}}
\end{tabular}
\label{tab:results}
\end{table}

\begin{figure}[!t]
    \centering
    \begin{subfigure}[b]{.5\linewidth}
        \centering
        \includegraphics[width=\linewidth]{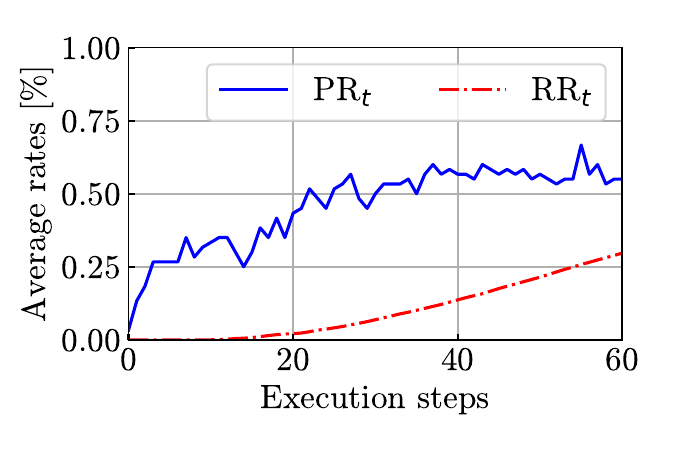}
        \caption{Mastermind}
        \label{fig:mastermind}
    \end{subfigure}
    \hspace{-.5em}
    \begin{subfigure}[b]{.5\linewidth}
        \centering
        \includegraphics[width=\linewidth]{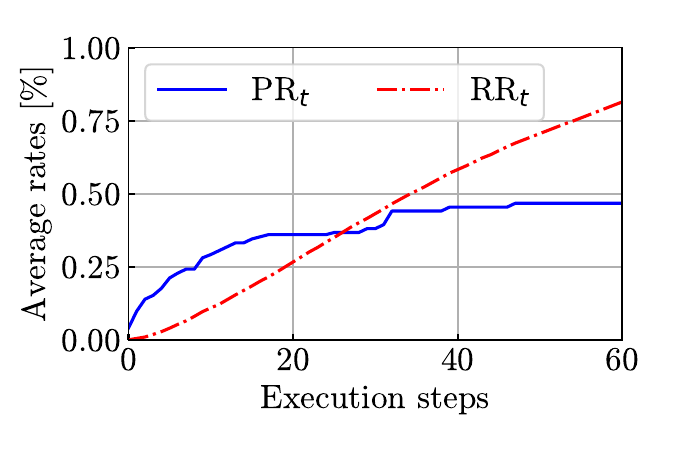}
        \caption{LTP}
        \label{fig:ltp}
    \end{subfigure}
    \caption{Average Progress rate PR$_t$ and the repetition rate RR$_t$ on Mastermind and LTP. Mastermind: It starts out with a low RR$_t$ but this increases after step 22 while the progress rate also stall at 55\%. LTP: at first a higher RR$_t$ allows the agent to progress by making small variations that lead to success, but later this plateaus.}
    \label{fig:trends}
\end{figure}

\begin{figure}[!t]
    \centering
    \begin{subfigure}[b]{.5\linewidth}
        \centering
        \includegraphics[width=\linewidth]{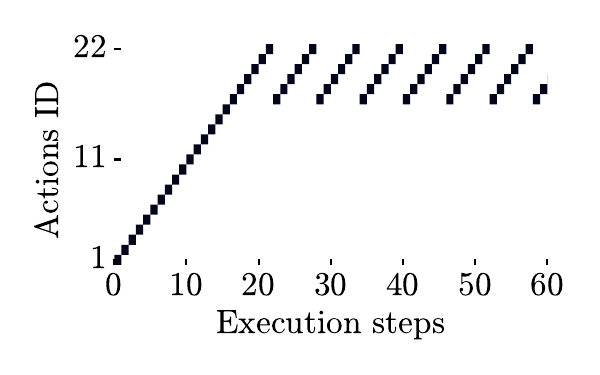}
        \caption{Mastermind}
        \label{fig:mastermind_loop}
    \end{subfigure}
    \hspace{-.5em}
    \begin{subfigure}[b]{.5\linewidth}
        \centering
        \includegraphics[width=\linewidth]{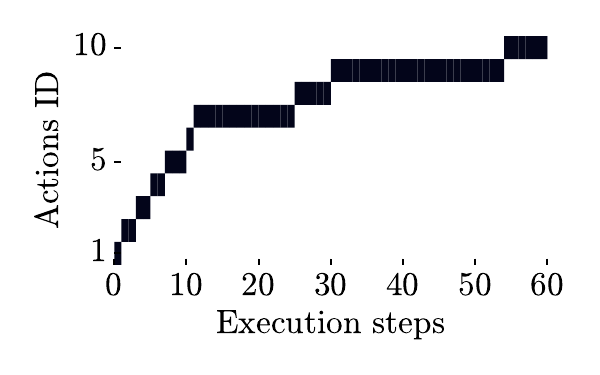}
        \caption{LTP}
        \label{fig:ltp_loops}
    \end{subfigure}
    \caption{Examples of repeated actions in Mastermind and LTP. Mastermind: there is a set of unique actions at first, but then gets stuck repeating the same actions over and over. LTP: repeated actions are small variations of the same question that lead to progress.}
    \label{fig:loops}
\end{figure}

\paragraph{Experimental results.} 
For Mastermind, \cref{fig:mastermind} shows the progress rate PR$_t$ and repetition rate RR$_t$. 
In the first 22 steps, the agent explores different solutions (RR$_{[0, 22]}<5\%$).  
This leads to growing progress towards the final goal, reaching half of the milestones (PR$_{22}\approx55\%$). Then, the agent starts performing the same actions, exhibiting a repetitive pattern (see also \cref{fig:mastermind_loop} rightmost part) and failing to reach the final goal within the next 38 steps. This results in a rise of the repetitions to RR$_{60}=30\%$ and a saturation of the progress rate at PR$_{60}=55\%$. Hence, \our offered us a crucial insights on why the current agent cannot solve the Mastermind game.

To overcome this agent limitation we incorporate a memory component~\cite{park2023generative} into the agent architecture. 
The agent stores the past guesses in a local buffer. Then, at each step, if the agent outputs an action already in the buffer, it is prompted to provide a new one. 
\cref{tab:results} (Mastermind$^*$) shows that this simple change in agent architecture has a big impact: the agent can now solve more instances, increasing the final SR from 47\% to 60\% and preventing repetitions (RR$_{60}=0\%$). This highlights how studying the interplay between progress and repetition rates can allow us to improve agent architecture, sometimes even with simple remedies. We support our intuition extending the evaluation to more instances of Mastermind from 15 to 60 achieving comparable results -- \ie 43\% of SR with the standard architecture and 62\% with the simple memory (19\% of improvement).

For LTP, the \our metrics reveal a different agent behaviour, where repetitions are part of the agent reasoning strategy, enhancing the progress rate (\cref{fig:ltp}).
From the initial steps, the agent changes aspects of the same questions until a local solution emerges. This leads to horizontal indicators in \cref{fig:ltp_loops} and RR$_{20}\approx30\%.$
Despite solving only a few riddles (SR=0.2), these repetitions contribute to progress, achieving 46\% of the milestones by the end of the execution, with a final repetition rate of RR$_{60}=81\%$. This shows us how the interplay of progress and repetition rates provides an insight on how agents behave across the different time steps.

Consider the benchmark ALFWorld in \cref{tab:results} (we report the metrics trend in Appendix \ref{s:app_alfworld_sudoku}). It requires the exploration of a textual world to locate an object. While the agent explores the solution space and limits action repetitions (RR$_{60}=6\%$), it fails to solve all the games (PR$_{60}=74\%$). This discrepancy may arise from the more exploration steps required to discover the object. We support this intuition extending the benchmark runtime to 120 steps resulting in a success and progress rates increase by 6\% (ALFWorld$^*$ in \cref{tab:results}). This confirms the usefulness of \our in understanding the agent failures.
We support our intuition also extending the evaluation to more instances of ALFWorld from 15 to 60 achieving comparable results -- \ie 83\% of SR with 60 steps as limit and 87\% with 120 steps as limit (4\% of improvement).

Finally, we look at Sudoku, known for its high level of difficulty~\cite{felgenhauer2006mathematics}. The low progress and repetition rates achieved after 60 steps (PR$_{60}=8\%$ and RR$_{60}=22\%$) indicate that the current agent architecture struggles in finding correct solutions solving this task. We report the metrics trend in Appendix \ref{s:app_alfworld_sudoku}.

%% file: sections/conclusions.tex
\section{Conclusions}
\label{s:conclusions}

\our allows the research community to keep track of agent progress in a holistic manner. Starting out with a first set of four benchmarks and two new metrics, \our is easily extendable. 
Furthermore, the two proposed metrics, progress and repetition rates, have the great advantage of allowing to track how agents advance toward the final goal over time. Especially studying their interplay can lead to important insights that will allow the research community to improve agent performance.
Finally, we believe that promptly sharing \our with the research community will facilitate benchmarking and debugging agents, and will foster the creation and use of new benchmarks and metrics.

\section*{Ethical Considerations}

The complexity of LLM agents poses challenges in comprehending their decision-making processes. Ethical guidelines must demand transparency in such systems, ensuring that developers and end-users comprehend how decisions are reached. 

We are not aware of any direct ethical impact generated by our work. However, we hope that insights into Generative AI agents' decision-making processes will be applied to improve and promote transparency and fairness.

%% file: appendix/app1.tex
\section{Appendix: ALFWorld and Sudoku benchmarks}
\label{s:app_alfworld_sudoku}

In this section we report the detailed metrics for each step for the ALFWorld and Sudoku benchmarks, omitted for the sake of brevity from the main paper.

\begin{figure}[!h]
    \centering
    \begin{subfigure}[b]{.5\linewidth}
        \centering
        \includegraphics[width=\linewidth]{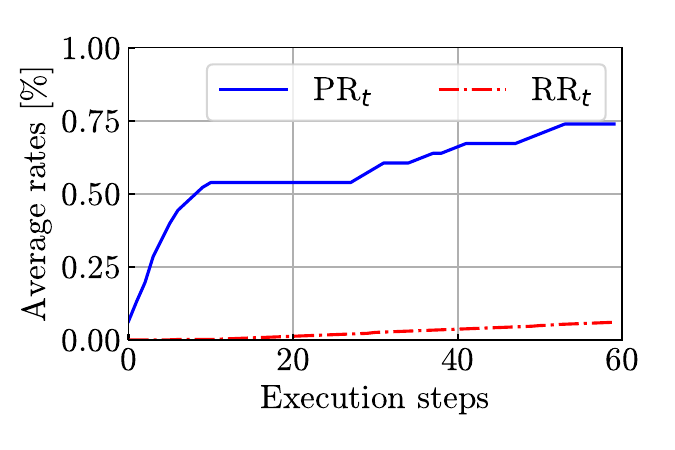}
        \caption{ALFWorld}
        \label{fig:alfworld}
    \end{subfigure}
    \hspace{-.5em}
    \begin{subfigure}[b]{.5\linewidth}
        \centering
        \includegraphics[width=\linewidth]{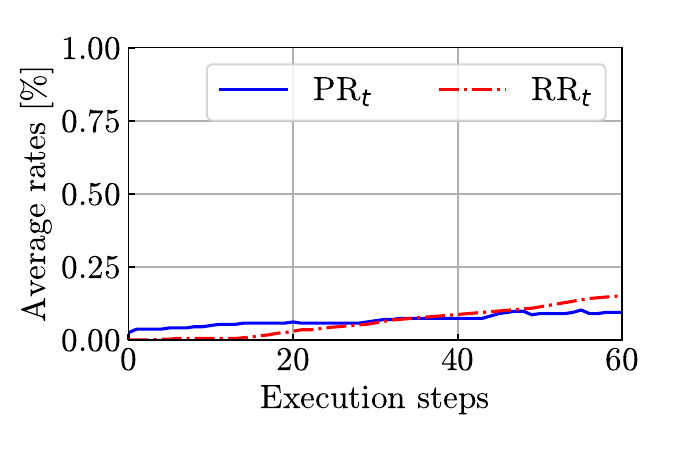}
        \caption{Sudoku}
        \label{fig:sudoku}
    \end{subfigure}
    \caption{Progress rate PR$_t$ and the repetition rate RR$_t$ on ALFWorld and Sudoku averaged over 15 runs. ALFWorld: It starts out with a low repetition rate and quick increase of the progress rate. Then a slow increase of the repetition rate enables to further increase the progress rate although less quickly. Sudoku: The progress rate quickly reaches 8\%. The repetition rate then slowly increases without any positive change in the progress rate.}
    \label{fig:trends_appendix}
\end{figure}

Figure~\ref{fig:alfworld} reports the progress rate and repetition rate for ALFWorld.
The repetition rate is close to 0\% for the first 20 steps, then it slowly increases up to 6\% after 60 steps.
The progress rate quickly reaches over 50\% in 10 steps, then keeps increasing, although slowly, up to 74\%.
The consistent improvement of the progress rate even for steps close to 60 together with the low repetition rate suggests that higher values may be reached by increasing the maximum number of steps.
We validate this hypothesis by extending the benchmark runtime to 120 steps.
As previously reported in Table \ref{tab:results}, this results in an improvement of 6 percentage points for both the success rate the progress rate, i.e. SR$=93\%$ and PR$_{120}=80\%$.

Figure~\ref{fig:sudoku} includes the two metrics for the Sudoku benchmark.
We can observe that the progress rate quickly reaches a plateau at 8\% in very few steps.
The repetition rate is close to 0\% for the first 10 steps, then it slowly increases up to 22\% after 60 steps without any improvement of the progress rate.

\section{Appendix: Additional agents architectures and benchmarks}
\label{sec:plugandplay}

In this section we highlight the plug-and-play aspect of \our showing the implementation of Mastermind with two additional agents architectures, \ie ReAct~\cite{yao2022react} as the most used architecture in literature and OpenAI Assistant~\cite{openai2023assistant}, as the most recent proprietary architecture.
Additionally, we show how to implement the open-ended benchmark GAIA~\cite{mialon2023gaia} requiring the usage of external tools.
For brevity, in the following snippets we omit details, like error handling or full agent definition. The complete code is available in the \href{https://github.com/nec-research/agentquest}{GitHub repository}.

\subsection{ReAct for Closed-box Environments}
We show an example of how to execute a closed-box benchmark (\ie ALFWorld) with an agent based on the ReAct architecture~\cite{yao2022react}. Such architecture forces the agent decision making process to generate both textual reasoning traces and actions pertaining to a task in an interleaved manner. Common implementations~\cite{chase2022langchain,yao2022react} rely on external tools to perform actions. Here, we ensure compatibility with existing implementations providing a single tool (\ie \texttt{ProxyTool}) that forwards the actions to the driver. In a nutshell, the agent reflects on the action to take and invokes the tool. Then, we feed the tool input to the driver to perform the interaction with the environment. At each step, we provide the agent the updated history of the actions and observations through the  \texttt{intermediate\_steps} variable.

\begin{python}
from agentquest.drivers import MasterMindDriver
from agentquest.metrics import ...
from agentquest.utils import Action
...

# Define a dummy tool for closed-box environments
class ProxyTool(BaseTool):
    name = "proxytool"
    description = "Provide the action you want to perform"
    def _run(self): 
        pass

# Instantiate custom prompt
prompt = CustomPromptTemplate(
    template=..., # LLM prompt
    tools=[ProxyTool()],
    input_variables=["intermediate_steps", ...]
)
# Initialise the agent
agent = create_react_agent(llm, [ProxyTool()], prompt)
intermediate_steps = []
# Initialise the driver
driver = MasterMindDriver(game)
# Get the first observation
obs = driver.reset()
# Agent Loop
while not obs.done:
    # Retrieve the agent output
    agent_choice = agent.invoke(
        {'input':obs.output,
         'intermediate_steps':intermediate_steps}
    )
    action = Action(action_value=agent_choice.tool_input)
    # Perform the step
    obs = driver.step(action)
    # Update intermediate steps
    intermediate_steps.append((agent_choice, obs.output))
    # Get current metrics ...
\end{python}

\subsection{OpenAI Assistant for Closed-box Environments}

The OpenAI Assistant~\cite{openai2023assistant} is a proprietary architecture. It allows users to define custom agents by specifying the tasks to accomplish and the set of tools the agent can use. While the decision-making process is not directly accessible by the end-users (the agent and the LLM are hosted on the proprietary cloud environment), the tools can be invoked both remotely or locally. In the latter, users have control on the tool invocation managing the agent loop. 

Similarly to ReAct, we here rely on the \texttt{ProxyTool}, acting as a proxy between the agent and the environment. We invoke the remote agent with the initial task (\eg first ALFWorld observation) and process the output of its decision making process, \ie the action to perform provided as tool input. Then, we bypass the tool invocation, directly forwarding the action to the driver to perform the execution step and retrieve the next observation. Finally, we invoke the agent with the new observation concluding the execution step.

\begin{python}
from agentquest.drivers import MasterMindDriver
from agentquest.metrics import ...
from agentquest.utils import Action
...

# Define a dummy tool for closed-box environments
class ProxyTool(BaseTool):
    name = "proxytool"
    description = "Provide the action you want to perform"
    def _run(self): 
        pass

# Initialise the agent
agent = OpenAIAssistantRunnable.create_assistant(
    instructions=... # LLM prompt
    tools=[ProxyTool()],
    model=... # Chosen LLM
    as_agent=True
)
# Initialise the driver
driver = MasterMindDriver(game)
# Get the first observation
obs = driver.reset()
# Get the first action
response = agent.invoke({"content": obs.output})
# Agent Loop
while not obs.done:
    # Retrieve the agent output
    agent_guess = response[0].tool_input
    action = Action(action_value=agent_guess)
    # Perform the step
    obs = driver.step(action)
    # Get current metrics ...
    # Manage Proxy Tool output
    tool_outputs = [
        {"output": obs.output, 
         "tool_call_id": response[0].tool_call_id}
    ]
    # Invoke the agent to get the next action
    response = agent.invoke(
        {"tool_outputs": tool_outputs,
         "run_id": response[0].run_id,
         "thread_id": response[0].thread_id}
    )
\end{python}

\subsection{OpenAI Assistant for Open-ended Environments}
When interacting with an open-ended environment, the agent is not restricted to the pre-defined actions of the closed-box environment and it is allowed to select any user-defined tool (\eg retrieving information online or executing code). Hence, we provide the agent the list of tools via the \texttt{tool} variable. The agent relies on its reasoning process to choose which tool to invoke. 
Omitted here for the sake of brevity, we rely of the manual annotations of the GAIA questions~\cite{mialon2023gaia} as milestones to compute the progress rate.

\begin{python}
from agentquest.drivers import GaiaDriver
from agentquest.metrics import ...
from agentquest.utils import Action
...

# Define the tools
tools=[
    OnlineSearch(), # Retrieve a web page link
    WebContentParser(), # Read the web page
    FinalAnswerRetriever(), # Provide the final answer
    ...
]
# Initialise the agent
agent = OpenAIAssistantRunnable.create_assistant(
    instructions=... # LLM prompt
    tools=tools,
    model=... # Chosen LLM
    as_agent=True
)
# Initialise the driver
driver = GaiaDriver(question, tools)
# Get the first observation
obs = driver.reset()
# Get the first action
response = agent.invoke({"content": obs.output})
# Agent Loop
while not obs.done:
    # Retrieve the agent output
    act = f'{response[0].tool}:{response[0].tool_input}'
    action = Action(action_value=act)
    # Perform the step invoking the local tool
    obs = driver.step(action)
    # Get current metrics ...
    # Manage tool output as observation
    tool_outputs = [
        {"output": obs.output, 
         "tool_call_id": response[0].tool_call_id}
    ]
    # Invoke the agent to get the next action
    response = agent.invoke(
        {"tool_outputs": tool_outputs,
         "run_id": response[0].run_id,
         "thread_id": response[0].thread_id}
    )
\end{python}
Here, the driver acts as a wrapper, executing the tool with the parameters provided by the agent (\texttt{tool\_input}) and forwards the output to the agent in the correct format:
\begin{python}
class GaiaDriver():
    def __init__(self, question, tools, ...):
        # Initialise the tool lookup
        self.tool_lookup = {x.name:x for x in tools}
    ...
    def step(self, action):
        # Parse the action
        tool, tool_input = action.action_value.split(':')
        # Invoke the tool
        tool_out = self.tool_lookup[tool]._run(tool_input)
        # Parse the tool output here ...
        return Observation(output=tool_out)
\end{python}